\documentclass[journal,10pt,twoside]{IEEEtran}
\usepackage{amsmath,amssymb}
\usepackage{graphicx}
\graphicspath{{figures/}}
\usepackage{booktabs,multirow,array,tabularx}
\usepackage[table]{xcolor}
\usepackage{tikz}
\usepackage{colortbl}
\usepackage{bm}
\usepackage{url}
\usepackage[nocompress]{cite}
\usepackage{microtype}
\usepackage[most]{tcolorbox}
\usepackage[hidelinks]{hyperref}
\usepackage[capitalise,nameinlink,noabbrev]{cleveref}
\definecolor{gainpositive}{RGB}{28,105,77}
\definecolor{gainnegative}{RGB}{166,56,52}
\definecolor{FEcontext}{HTML}{2EA69A}
\definecolor{FEentity}{HTML}{8847E9}
\newtcolorbox{FEpromptbox}{%
  enhanced,
  width=0.96\columnwidth,
  colback=white,
  colframe=black!60,
  colbacktitle=black!78,
  coltitle=white,
  fonttitle=\small\bfseries,
  title={MME-RealWorld-RS Prompt},
  boxrule=0.5pt,
  arc=1pt,
  left=9pt,right=9pt,top=7pt,bottom=8pt,
  before skip=5pt,after skip=5pt
}

\newcommand{\figref}[1]{\hyperref[#1]{Fig.~\ref*{#1}}}
\newcommand{\tabref}[1]{\hyperref[#1]{Table~\ref*{#1}}}
\newcommand{\cmark}{\checkmark}
\newcommand{\xmark}{\times}
\DeclareMathOperator*{\argmin}{arg\,min}
\DeclareMathOperator*{\argmax}{arg\,max}
\providecommand{\FENa}{--}
\title{FocusEarth: Where to Look and What to Preserve for Ultra-High-Resolution Remote Sensing VQA}
\author{Wanghao~Sun, Nanqing~Liu, Yu~Sun, Zhanxuan~Hu, Qi~Ming, and Heng-Chao~Li%
\thanks{This work was supported in part by the National Natural Science Foundation of China under Grant 62601704. (Corresponding author: Nanqing Liu.)}%
\thanks{Wanghao Sun, Nanqing Liu, Yu Sun, and Zhanxuan Hu are with the School of Information Science and Technology, Yunnan Normal University, Kunming, China (e-mail: lansing163@163.com, zhanxuanhu@gmail.com).}%
\thanks{Qi Ming is with the College of Computer Science, Beijing University of Technology, Beijing, China (e-mail: chaser.ming@gmail.com).}%
\thanks{Heng-Chao Li is with the School of Information Science and Technology, Southwest Jiaotong University, Chengdu 611756, China (e-mail: lihengchao\_78@163.com).}}
\begin{document}
\bstctlcite{IEEEexample:BSTcontrol}
\maketitle

\begin{abstract}
Ultra-high-resolution (UHR) remote sensing visual
question answering (RS-VQA) requires interpreting
fine-grained local details within their global
spatial context under limited visual-token budgets.
Region selection identifies where relevant evidence
may lie, but does not determine which encoded features
to retain when compressing the full scene.
We propose FocusEarth, a training-free framework
that couples
\emph{where to look} in the source image with
\emph{what to preserve} in the encoded visual
representation.
Entity and context descriptions derived from each question
capture the objects of interest and their spatial
relationships. Multiple frozen
image--text encoders score full-region and local views
of candidate source regions. Their scores are
standardized and their regional distributions matched
in entropy before fusion, allowing complementary
responses to guide selection under a fixed region
budget. The selected source regions are then mapped
onto the encoded feature grid of a resized full-scene
image. Their corresponding features are retained
individually, while the remaining features are
spatially aggregated,
preserving selected evidence
within a compact, spatially aligned representation.
Across four UHR RS-VQA benchmarks, the method ranks
first in overall accuracy on three. On MME-RealWorld-RS,
it improves direct Qwen3-VL-8B inference by 13.80
percentage points while reducing visual-token count
by 54.0\% and answering-stage latency by 49.0\%,
with region scoring and selection precomputed offline.
The code is available at
\url{https://github.com/sun-wanghao/FocusEarth}.
\end{abstract}


\section{Introduction}
\label{sec:introduction}

Remote sensing visual question answering (RS-VQA) enables
natural-language interaction with aerial and satellite imagery,
supporting object recognition, counting, and spatial
reasoning~\cite{lobry2020rsvqa,wang2024earthvqa}.
Recent multimodal large language models (MLLMs), such as
GeoChat~\cite{kuckreja2024geochat} and
EarthGPT~\cite{zhang2024earthgpt}, have advanced remote sensing
image understanding.
However, ultra-high-resolution (UHR) imagery presents particular
challenges due to its large spatial extent, substantial variation
in object scales, and visually similar instances.
Answering questions about such scenes often requires interpreting
fine-grained local details within their global spatial context.
The disparity between image resolution and practical model input
capacity makes it difficult to accommodate both within a limited
computational budget~\cite{zhang2025mmerealworld,wang2025xlrsbench}.

Recent studies have addressed UHR RS-VQA through high-resolution
model adaptation~\cite{wang2025geollava8k}, text-guided visual
token reduction~\cite{luo2025largersi}, and learned active
perception~\cite{liu2026zoomearth,wang2026beyondzooming}.
These approaches rely on additional model training, motivating
the exploration of inference-time strategies for frozen MLLMs.
Among training-free approaches, ZoomSearch~\cite{zhou2025zoomsearch}
uses adaptive region search to locate question-relevant content
and reassembles the selected patches,
WeaveEarth~\cite{ma2026weaveearth} constructs structured visual
evidence, and GazeEarth~\cite{zhang2026gazeearth} adopts
full-scene foveated mapping.
These methods differ in how they select and organize visual
evidence, but leave a related question: once relevant source
regions are identified, how should their locations determine
which features in a full-scene representation are retained?
Answering this requires considering both which regions matter
to the query and how their evidence can be preserved alongside
the global scene context.

\begin{figure*}[!t]
    \centering
    \includegraphics[width=0.98\textwidth]{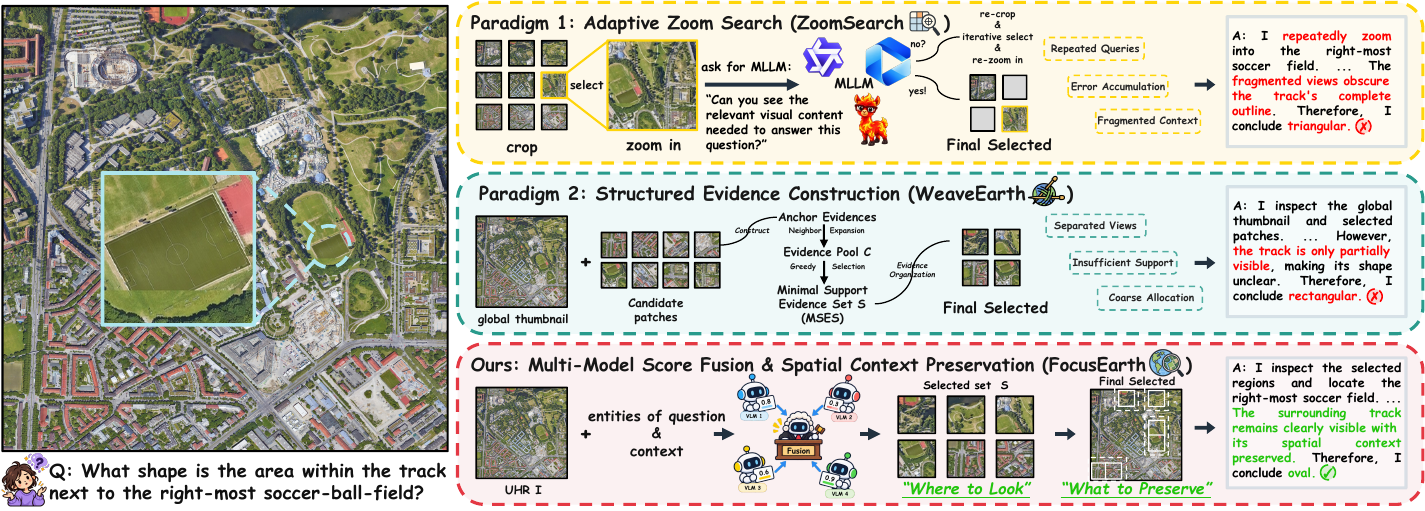}
    \caption{Motivation for using question-relevant regions to guide
    feature-level compression in UHR RS-VQA, illustrated by a query
    about the shape of the area inside the track next to the
    rightmost soccer field. ZoomSearch~\cite{zhou2025zoomsearch}
    employs adaptive region search and patch reassembly, while
    WeaveEarth~\cite{ma2026weaveearth} organizes local evidence
    alongside a global thumbnail. FocusEarth uses selected regions
    to preserve corresponding encoded features within a single
    resized full-scene image.}
    \label{fig:motivation}
\end{figure*}

\textbf{Where to look.}
Localizing question-relevant regions in UHR imagery requires
considering both the entities mentioned in a question and
their spatial context. As illustrated in \figref{fig:motivation},
identifying the intended track involves locating both the
track and the soccer field while accounting for their spatial
relationships. Entity cues facilitate object localization,
whereas contextual cues help distinguish visually similar
instances. Furthermore, different image--text
encoders~\cite{radford2021clip,liu2024remoteclip,
zhang2024rs5m,wang2024skyscript} can provide complementary
regional responses to the same textual query.
However, variations in their score distributions complicate
direct fusion.

\textbf{What to preserve.}
Once relevant regions have been localized, their visual
evidence needs to be interpreted in relation to the surrounding
scene. Existing image-level approaches accommodate local
details and global context through different spatial
arrangements~\cite{zhou2025zoomsearch,ma2026weaveearth}
or nonuniform resolution allocation~\cite{zhang2026gazeearth}.
Retaining the complete scene at sufficient resolution,
however, can result in long visual token sequences.
Visual token reduction provides a complementary means of
controlling computational
cost~\cite{bolya2023tome,chen2024fastv,yang2025visionzip,
shang2025prumerge}.
The spatial correspondence between localized source regions
and encoded visual features provides a basis for determining
which features to retain during compression.
The challenge is therefore to retain question-relevant
features within a compact representation while maintaining
the full-scene spatial layout.

We develop \textbf{FocusEarth}, a training-free framework
that uses source-region localization to guide visual token compression.
For region selection, FocusEarth derives entity and context
descriptions from each question and ranks candidate regions
using multiple frozen image--text encoders, whose regional
responses are fused after score standardization and entropy
matching. The resulting entity and context rankings jointly
guide region selection.
FocusEarth maps the selected regions onto the encoded feature
grid of a resized full-scene image, retaining their corresponding
features individually while aggregating the remaining features
within spatial neighborhoods.
A frozen MLLM answers the original question using the resulting
compact visual representation.

The main contributions of this work are summarized as follows:
\begin{itemize}
  \item We propose \textbf{FocusEarth}, a training-free framework for
  UHR RS-VQA that maps question-relevant source regions to
  encoded full-scene features to guide selective token preservation
  and spatial compression.

  \item We introduce a \textbf{Query-Adaptive Region Selection} strategy
  that leverages multiple image--text encoders to select regions
  based on entity and contextual relevance.

  \item We construct a \textbf{Spatially Aligned Visual Representation}
  by retaining selected-region features and compressing background
  tokens while maintaining the full-scene spatial layout.

  \item We demonstrate the effectiveness of FocusEarth on four
  UHR RS-VQA benchmarks. Evaluations across frozen answer models
  show accuracy gains, while efficiency experiments reveal substantial
  reductions in visual tokens and answering-stage latency.
\end{itemize} 

\section{Related Work}
\label{sec:related_work}

\subsection{Remote Sensing Vision--Language Models}

Early studies on RS-VQA established benchmarks and developed
task-specific models for object recognition, attribute
identification, and spatial reasoning~\cite{lobry2020rsvqa}.
SHRNet~\cite{zhang2023shrnet} enhances spatial reasoning through
hierarchical representations, while EarthVQA~\cite{wang2024earthvqa}
emphasizes object-centric relational reasoning.
With the development of pretrained multimodal models,
RSAdapter~\cite{wang2024rsadapter} explores their adaptation
to remote sensing VQA.
More recent MLLMs extend remote sensing image understanding
to grounded dialogue and multisensor interpretation, as
exemplified by GeoChat~\cite{kuckreja2024geochat} and
EarthGPT~\cite{zhang2024earthgpt}.
Benchmarks such as VRSBench~\cite{li2024vrsbench} and
CHOICE~\cite{an2025choice} further broaden the evaluation
of remote sensing vision--language understanding.
Foundation vision models have also been adapted for remote
sensing dense prediction~\cite{ma2026evsr,liu2025pointsam}.

Alongside generative vision--language models, contrastive
image--text encoders provide a basis for estimating semantic
relevance between visual content and textual descriptions.
CLIP~\cite{radford2021clip} establishes general-purpose
image--text alignment, while
RemoteCLIP~\cite{liu2024remoteclip},
GeoRSCLIP~\cite{zhang2024rs5m}, and
SkyCLIP~\cite{wang2024skyscript} extend contrastive
vision--language learning to remote sensing imagery.
RegionCLIP~\cite{zhong2022regionclip} further explores
language--image alignment at the region level.
Recent studies also investigate the fusion of pretrained
model predictions to improve zero-shot
recognition~\cite{lu2024beyond,hu2026sota}.
In contrast to class-level prediction fusion, FocusEarth
combines regional relevance responses from multiple
image--text encoders under entity and context descriptions
for question-guided region selection.

\begin{figure*}[!t]
    \centering
    \includegraphics[width=0.92\linewidth]{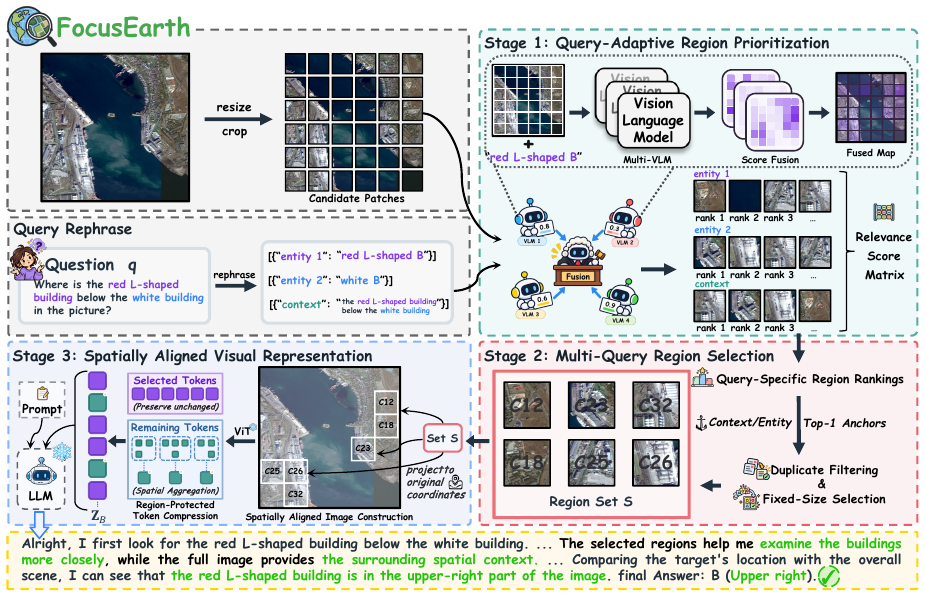}
    \caption{Overview of FocusEarth. Entity and context descriptions
    guide region ranking and selection through multi-encoder fusion.
    The selected source regions are mapped onto the encoded feature
    grid of a resized full-scene image. Their corresponding features
    are retained individually, while background features are
    spatially aggregated. A frozen MLLM answers the original question
    using the resulting compact visual representation.}
    \label{fig:overview}
\end{figure*}

\subsection{UHR Remote Sensing Image Understanding}

High-resolution vision--language models accommodate large
images through specialized visual encoding and token reduction.
LLaVA-UHD~\cite{guo2024llavauhd} combines image slicing with
visual token compression, while
GeoLLaVA-8K~\cite{wang2025geollava8k} introduces background
token pruning and anchored token selection for UHR remote
sensing imagery.
Coarse-to-fine text-guided pruning~\cite{luo2025largersi}
further exploits question relevance to select informative
regions across multiple scales.
ImageRAG~\cite{zhang2025imagerag} formulates UHR image
understanding as a retrieval problem over long visual contexts.
Active perception provides another means of acquiring
task-relevant visual evidence through adaptive
zooming~\cite{liu2026zoomearth,wang2026geoeyes},
multi-tool reasoning~\cite{wang2026beyondzooming}, and
iterative visual exploration~\cite{zhang2026geovis,zhu2026geovista}.

Recent training-free approaches further investigate how
localized visual evidence can be organized for question
answering.
ZoomSearch~\cite{zhou2025zoomsearch} combines hierarchical
multi-branch search with layout-aware patch reassembly.
WeaveEarth~\cite{ma2026weaveearth} integrates global-aware
evidence selection with a structured representation comprising
a global thumbnail, local evidence, and spatial metadata.
GazeEarth~\cite{zhang2026gazeearth} employs topology-preserving
foveated mapping to redistribute spatial resolution within
a continuous scene.
While these methods organize visual evidence at the image level,
FocusEarth extends source-region selection to feature-level
compression by establishing spatial correspondence between
the selected regions and encoded visual features.

\subsection{Visual Token Reduction}

Visual token reduction has been widely explored to improve
the computational efficiency of multimodal models.
ToMe~\cite{bolya2023tome} merges similar tokens in vision
transformers, whereas FastV~\cite{chen2024fastv} prunes
visual tokens within the language model.
VisionZip~\cite{yang2025visionzip} and
PruMerge~\cite{shang2025prumerge} combine informative
token selection with feature aggregation.
Beyond visual redundancy, DivPrune~\cite{alvar2025divprune}
selects tokens based on feature diversity, while
CDPruner~\cite{zhang2025cdpruner} incorporates instruction
relevance into diversity-based selection.
SparseVLM~\cite{zhang2025sparsevlm} and
QuietPrune~\cite{gao2026quietprune} further exploit textual
guidance for visual token pruning.

For high-resolution imagery, Pyramid Token
Pruning~\cite{liang2026ptp} integrates region-level saliency,
token importance, and instruction relevance.
In remote sensing, coarse-to-fine text-guided
pruning~\cite{luo2025largersi} combines region localization
with multiscale visual token pruning.
UHR-BAT~\cite{dang2026uhrbat} incorporates query-guided
regional importance into budget-aware token compression,
while DualComp~\cite{li2026dualcomp} employs task-adaptive
semantic aggregation and geometric structure recovery.
FocusEarth instead derives region-level protection from
multiple image--text encoders operating on the source image
and maps the selected regions onto the encoded feature grid,
retaining their corresponding features individually while
spatially aggregating the remaining features.

\section{Methodology}
\label{sec:methodology}

\subsection{Framework Overview}
\label{sec:problem_overview}

Given a UHR remote sensing image $I$ and a question $q$,
FocusEarth first identifies source regions relevant to
the question. It then maps these regions onto the encoded
feature grid of a single resized full-scene image.
Features at the selected locations are kept as individual
tokens, while the remaining features are spatially
aggregated. In this way, region selection determines
which visual evidence survives compression without
rearranging the scene. As illustrated in
\figref{fig:overview}, the framework is summarized by
\begin{align}
\mathcal{S} &= \operatorname{Select}_{K}\!\left(
  \operatorname{Rank}(I,\operatorname{Describe}(q))\right),
  \nonumber\\
\hat{a} &= \operatorname{Answer}_{\theta}\!\left(
  q,\operatorname{Visual}(I,\mathcal{S})\right).
\label{eq:framework}
\end{align}
Here, $\mathcal{S}$ contains the selected source-region
indices under a region budget $K$.
$\operatorname{Visual}$ constructs the compact visual
representation from $I$ and $\mathcal{S}$, and a frozen
MLLM with parameters $\theta$ predicts the answer
$\hat{a}$ from that representation and the original
question, including answer choices when applicable.

\textbf{Stage 1: Query-Adaptive Region Prioritization.}
Deterministic parsing extracts entity and context
descriptions from the question to capture object cues
and spatial constraints. Multiple frozen image--text
encoders score full-region and local views of each
candidate source region. Calibrating and fusing their
responses produces separate rankings for the entity
and context descriptions.

\textbf{Stage 2: Multi-Query Region Selection.}
The top context region and the leading region for each
entity description are proposed in that order under a
shared budget $K$. After duplicate proposals are removed,
the context ranking fills any remaining slots. This
combines scene-level constraints with coverage of the
objects named in the question.

\textbf{Stage 3: Spatially Aligned Visual Representation.}
Selected regions are outlined on the resized full-scene
image and mapped onto its encoded feature grid. Their
corresponding features remain individual tokens; the
other features are grouped and aggregated under a
visual-token budget. The resulting sequence retains
the scene coordinates of selected features and spatial
representatives of the background groups for the
frozen answer model.

\begin{figure}[!t]
\centering
\includegraphics[width=0.96\columnwidth]{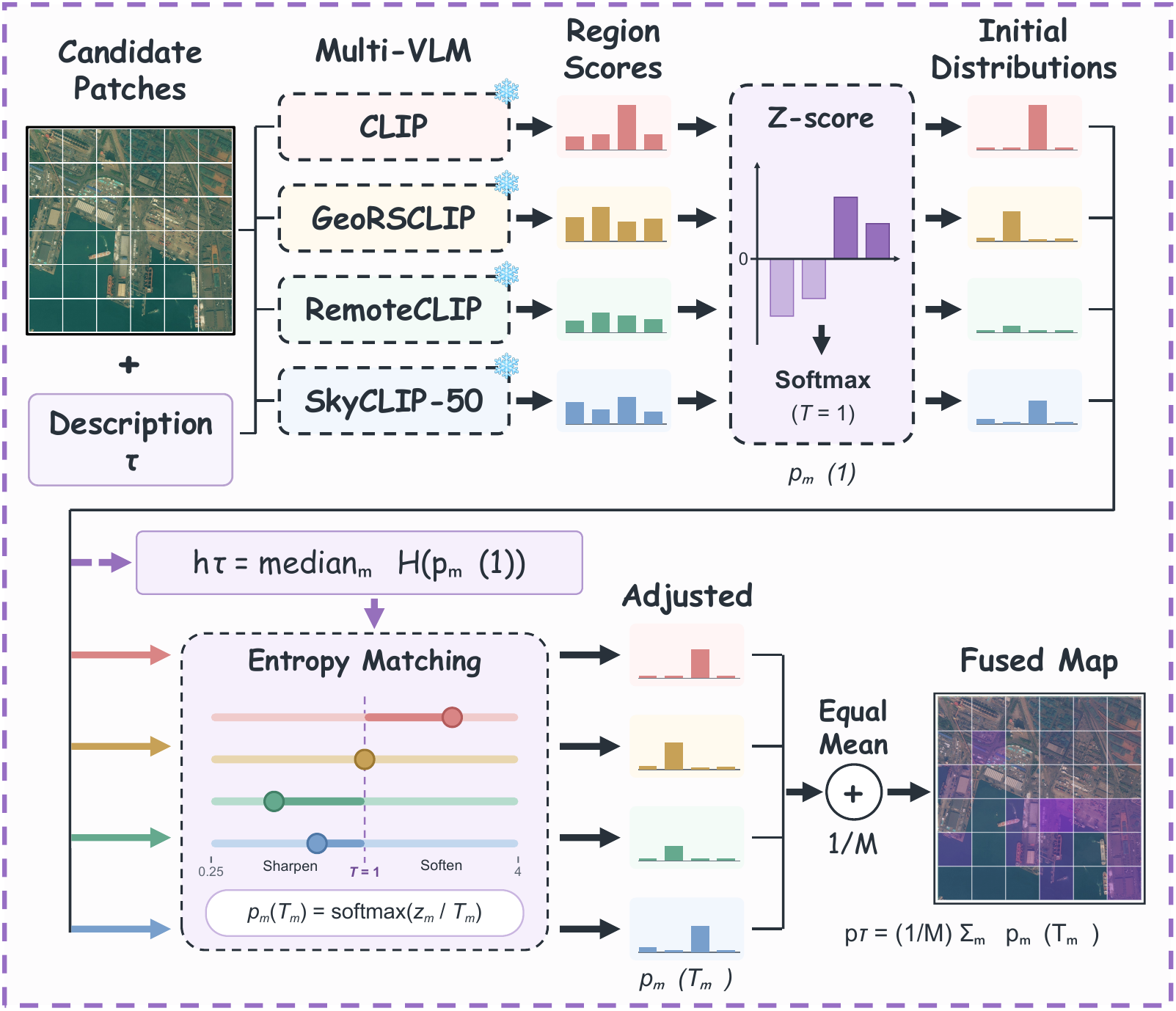}
\caption{Query-Adaptive Region Prioritization for one description $\tau$. Encoder scores are standardized and their regional distributions are matched in entropy before equal-weight fusion.}
\label{fig:region_prioritization}
\vspace{-2mm}
\end{figure}

\subsection{Query-Adaptive Region Prioritization}
\label{sec:region_prioritization}

Region prioritization considers both the objects mentioned
in a question and the spatial context that identifies the
intended instance. We derive entity descriptions and a
context description through deterministic parsing with the
spaCy~\cite{spacy} \texttt{en\_core\_web\_sm} parser.
Entity descriptions retain nominal heads and identity
modifiers, with a target-recovery rule incorporating
explicitly queried nouns missed during extraction.
The context description removes interrogative phrasing
and whole-image boilerplate while retaining object cues,
relations, negation, cardinality, and instance selectors.
Entity descriptions preserve their occurrence order,
including repeated mentions, while empty or placeholder
entities are excluded. If context reformulation yields
an empty description, the original question is used
for context ranking.

Following the image--text region-scoring approaches of
ZoomSearch~\cite{zhou2025zoomsearch} and
WeaveEarth~\cite{ma2026weaveearth}, we estimate regional
relevance for each entity description
$\tau_1,\ldots,\tau_J$ and the context description
$\tau_c$ over a common candidate set
$\mathcal{P}=\{P_i\}_{i=1}^{N}$.
The candidates form a fixed row-major grid on the
source image. Each candidate is represented by its
full-region view and local corner crops extracted
before encoder-specific preprocessing. The full-region
view captures regional context, while the local views
provide finer observations of its contents.
To account for variations in regional responses
across image--text encoders, we combine the scores
from $M$ frozen encoders~\cite{radford2021clip,
liu2024remoteclip,zhang2024rs5m,wang2024skyscript}
for each description, as illustrated in
\figref{fig:region_prioritization}.

For clarity, we omit the explicit dependence on $\tau$
in the following formulation.
Encoder $m$ assigns candidate $i$ its maximum
image--text similarity across the associated views:
\begin{equation}
s_{m,i}=\max_{v\in\mathcal{V}_i}
\left\langle f_m(v),g_m(\tau)\right\rangle.
\label{eq:region_view_score}
\end{equation}
Here, $\mathcal{V}_i$ denotes the set of full-region
and local views associated with candidate $i$,
while $f_m$ and $g_m$ are the image and text branches
of encoder $m$, respectively, with $\ell_2$-normalized
outputs.
To reduce differences in score location and scale
across encoders, we standardize the scores over
the $N$ candidates as
$z_{m,i}=(s_{m,i}-\mu_m)/\max(\sigma_m,\epsilon_s)$,
where $\mu_m$ and $\sigma_m$ are the mean and population
standard deviation of the candidate scores, respectively,
and $\epsilon_s>0$ prevents division by zero.
Temperature scaling~\cite{guo2017calibration} then
controls the concentration of the regional relevance
distribution without changing the candidate ranking:
\begin{equation}
p_{m,i}(T) =
\frac{\exp(z_{m,i}/T)}
{\sum_{j=1}^{N}\exp(z_{m,j}/T)}.
\label{eq:region_distribution}
\end{equation}

Although score standardization reduces scale
differences, the resulting distributions may still
vary in concentration. More concentrated distributions
can exert a disproportionate influence on the fused
regional responses despite equal encoder weights.
We therefore perform entropy matching before fusion.
For description $\tau$, the target entropy is defined
as the median of the unit-temperature entropies across
encoders,
$h_\tau=\operatorname{median}_{m}
\mathcal{H}(\mathbf{p}_m(1))$,
where
$\mathcal{H}(\mathbf{p})=-\sum_i p_i\log p_i$.
The temperature of each encoder is determined within
a bounded positive interval by minimizing its deviation
from the target entropy:
\begin{equation}
T_m = \argmin_{T\in[T_{\min},T_{\max}]}
\left|\mathcal{H}(\mathbf{p}_m(T))-h_\tau\right|.
\label{eq:entropy_matching}
\end{equation}
The resulting distributions are fused by equal-weight
averaging as
$\mathbf{p}_\tau=M^{-1}\sum_{m=1}^{M}
\mathbf{p}_m(T_m)$.
The fused entity and context rankings provide the
input to Multi-Query Region Selection.

\subsection{Multi-Query Region Selection}
\label{sec:multi_query_selection}

Context and entity rankings capture different aspects of
question relevance. Selecting regions solely from the
context ranking may overlook individual objects mentioned
in the question. We therefore integrate context and entity
proposals under a shared region budget $K$, where
$1\leq K\leq N$ (\figref{fig:overview}).
For each description $\tau$, the top-ranked candidate is
$a_\tau=\argmax_{1\leq i\leq N}p_{\tau,i}$.
Candidate-index order resolves score ties.
The region proposals are arranged as an ordered sequence,
with the context proposal followed by the entity proposals:
$\mathcal{R}=(a_{\tau_c},a_{\tau_1},\ldots,a_{\tau_J})$.
This ordering prioritizes the contextual description while
incorporating individual object cues.
We retain the first $K$ distinct candidate indices in
$\mathcal{R}$, or all distinct indices if fewer than $K$
are available, to form the proposal set $\mathcal{A}$.

The remaining budget is allocated according to the context
ranking, excluding candidates already included in
$\mathcal{A}$. The final region set is defined as
\begin{equation}
\mathcal{S}=\mathcal{A}\cup
\operatorname{Top}_{K-|\mathcal{A}|}
\bigl(\mathbf{p}_{\tau_c},\,\{1,\ldots,N\}\setminus\mathcal{A}\bigr),
\label{eq:region_selection}
\end{equation}
where $\operatorname{Top}_{r}(\mathbf{p},\mathcal{U})$
returns the $r$ highest-scoring candidate indices in
$\mathcal{U}$ according to $\mathbf{p}$, or an empty set
when $r=0$. This formulation yields $|\mathcal{S}|=K$.
In the absence of entity proposals, region selection
reduces to context-based top-$K$ selection.
The corresponding source boxes are ordered by their
spatial positions and used to guide image construction
and feature protection.

\subsection{Spatially Aligned Visual Representation}
\label{sec:visual_representation}

\begin{figure}[!t]
\centering
\includegraphics[width=\columnwidth]{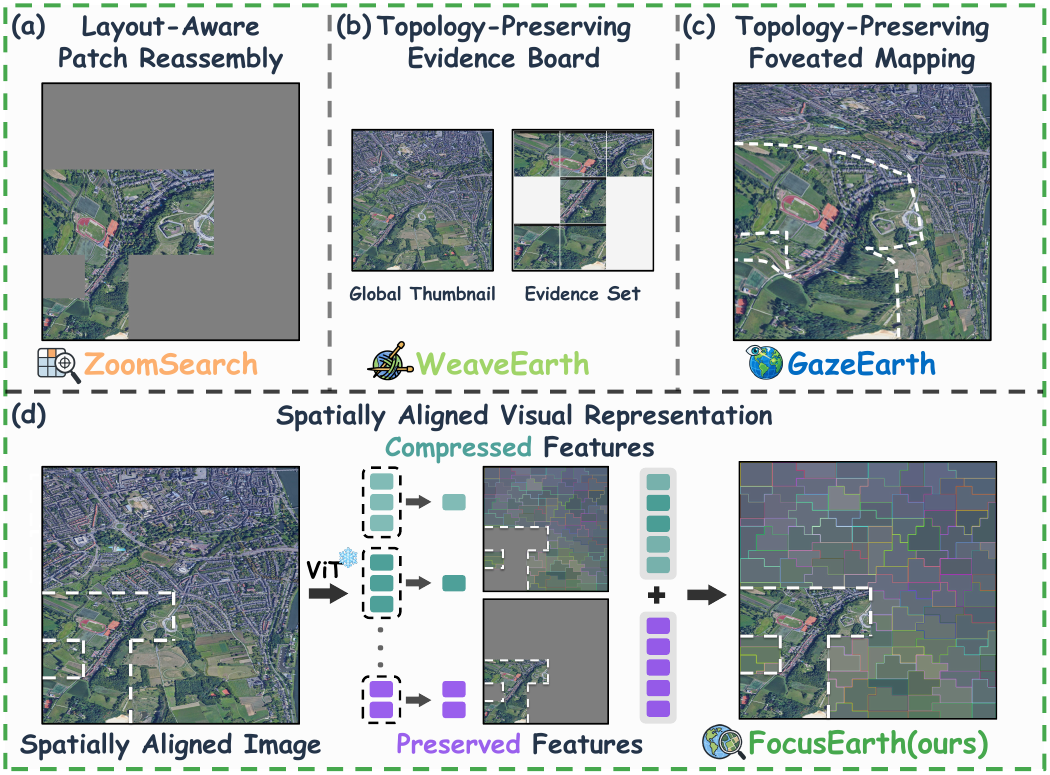}
\caption{Comparison of visual evidence representations
for UHR RS-VQA: (a) layout-aware patch reassembly
in ZoomSearch, (b) structured evidence organization
in WeaveEarth, (c) full-scene foveated mapping in
GazeEarth, and (d) spatially aligned full-scene
representation in FocusEarth.}
\label{fig:visual_representation}
\vspace{-0.4cm}
\end{figure}

Once question-relevant regions have been selected,
the visual representation needs to accommodate their
local features and the surrounding scene context within
a limited token budget.
\figref{fig:visual_representation} illustrates different
approaches to organizing visual evidence.
ZoomSearch~\cite{zhou2025zoomsearch} reassembles
selected patches into a compact canvas while retaining
their relative arrangement and coarse global positions
[Fig.~\ref{fig:visual_representation}(a)].
WeaveEarth~\cite{ma2026weaveearth} organizes local
evidence in a topology-preserving board, supplemented
by a global thumbnail and spatial metadata
[Fig.~\ref{fig:visual_representation}(b)].
GazeEarth~\cite{zhang2026gazeearth} reallocates the
pixel budget through topology-preserving foveated
mapping, enlarging the selected neighborhood while
compressing the surrounding scene
[Fig.~\ref{fig:visual_representation}(c)].
FocusEarth maintains the source-relative spatial
layout in a single resized full-scene image
[Fig.~\ref{fig:visual_representation}(d)] and extends
region selection to feature-level compression.
By mapping the selected source regions onto the
encoded feature grid, FocusEarth retains their
corresponding features individually and spatially
aggregates the remaining features.

\subsubsection{Spatially Aligned Image Construction}
\label{sec:canvas}

To establish spatial correspondence between selected
source regions and encoded visual features, FocusEarth
represents the complete scene as a single resized image.
A reference canvas is determined by the source aspect
ratio and selected-box coverage, with its dimensions
adjusted to those compatible with the vision processor.
The full RGB image is bilinearly resized to this canvas,
and the selected source boxes are mapped to their
corresponding spatial locations.
Outlines are rendered outside the union of the selected
boxes, leaving their interiors unobscured.
The outlines provide explicit visual cues to the
selected locations, whereas a corresponding feature-level
protection mask governs token compression.
Source crops are used only for region scoring, while
the resized full-scene image serves as the visual input
to the answer model.

Let $\widetilde{P}_i$ denote the mapped box of selected
region $i$, expanded to include its outer outline and
clipped to the image bounds.
After visual encoding and initial token merging,
the resulting feature grid contains $n_r$ rows and
$n_c$ columns, with a total of $L=n_rn_c$ cells.
Let $\mathcal{C}_j$ denote the image area corresponding
to feature cell $j$. Both $\mathcal{C}_j$ and
$\widetilde{P}_i$ are represented as half-open
rectangular regions.
The protected feature indices are defined as
\begin{equation}
\mathcal{I}_{\mathrm{sel}} =
\bigl\{j\in\{1,\ldots,L\}:\exists i\in\mathcal{S},\,
\mathcal{C}_j\cap\widetilde{P}_i\neq\varnothing\bigr\}.
\label{eq:region_token_mapping}
\end{equation}
For a clipped box envelope $[x_1,y_1,x_2,y_2)$ on
a $W_c\times H_c$ canvas, the corresponding zero-based
feature-grid column indices range from
$\lfloor x_1n_c/W_c\rfloor$ through
$\lceil x_2n_c/W_c\rceil-1$.
Row indices are determined analogously using $H_c$
and $n_r$.
The union of these cells over all selected regions
defines the protection mask, which identifies the
features to be retained individually during
subsequent compression.

\subsubsection{Region-Protected Token Compression}
\label{sec:spatial_aggregation}

The spatial correspondence established above allows
selected and unselected regions to be treated differently
during visual token compression.
Protected features are retained individually, while
the remaining features are spatially aggregated to
provide a compact representation of the surrounding scene.

For each question $q$, let $L_q$ denote the post-merger
visual-grid length, $R_q=|\mathcal{I}_{\mathrm{sel}}|$
the number of protected cells, and $C_q$ the number
of four-neighbor connected components formed by
unprotected cells.
Let $p_q^{\mathrm{std}}$ denote the maximum protected-cell
cost of a fully represented standard candidate region.
We determine the nominal protection reserve $E_q$,
the base allocation for unprotected context
$O_q^{\mathrm{base}}$, and the final visual-token
budget $B_q$ as
\begin{equation}
\begin{aligned}
E_q &= Kp_q^{\mathrm{std}}, \\
O_q^{\mathrm{base}} &= \max\!\left(C_q,\left\lceil
\rho(L_q-E_q)\right\rceil\right),\\
B_q &= \min\!\left(L_q,E_q+O_q^{\mathrm{base}}\right).
\end{aligned}
\label{eq:token_budget}
\end{equation}
Here, $\rho\in[0,1]$ controls the nominal allocation
for unprotected context and is set to $0.605$ in the
default configuration.
Under the standard region setting, $R_q\leq E_q$,
which yields $C_q\leq B_q-R_q\leq L_q-R_q$.
The protected cells occupy $R_q$ output tokens,
while the remaining $B_q-R_q$ tokens represent
unprotected context, including any unused protection
reserve.

To preserve spatial locality, aggregation is restricted
to four-neighbor connected components of the
unprotected feature grid.
The available context tokens are allocated across
these components in proportion to their sizes using
deterministic integer rounding, with at least one
group assigned to each component.
Within each component, seeds are selected at quantiles
of the Morton-ordered cells.
Multi-source breadth-first expansion from these seeds
partitions the component into spatially connected groups.

Together, the protected cells and context groups
partition the post-merger grid into $B_q$ groups
$G_1,\ldots,G_{B_q}$, with each protected cell forming
a singleton group.
Let $\mathbf{z}_j\in\mathbb{R}^{D}$ denote the
post-merger feature of cell $j$ and
$\mathbf{x}_j\in\mathbb{Z}^{2}$ its original row--column
coordinate.
Each group is represented by its mean feature and
the coordinate of the member $j_b\in G_b$ closest
to its coordinate centroid, with row-major order
resolving ties:
\begin{equation}
\widehat{\mathbf{z}}_b =
\frac{1}{|G_b|}\sum_{j\in G_b}\mathbf{z}_j,
\qquad
\widehat{\mathbf{x}}_b = \mathbf{x}_{j_b}.
\label{eq:group_aggregation}
\end{equation}
This formulation retains the original features and
coordinates of protected cells while representing
unprotected groups through spatially localized
feature summaries.
The output tokens are ordered by their representative
coordinates in row-major order.

For Qwen3-VL~\cite{bai2025qwen3vl}, the same groups
and representative coordinates are applied to the
final visual features and DeepStack streams.
The image-placeholder sequence is shortened to
match the retained visual length, while the visual
encoder processes the original image input and
grid metadata.
Visual position encoding follows the representative
source-grid coordinates, with text positions and
rotary offsets adjusted accordingly.
The resulting compact visual sequence is provided
to the frozen answer model together with the
original question.

\section{Experiments}
\label{sec:experiments}

\subsection{Experimental Setup}
\label{sec:experimental_setup}

\noindent\textbf{Benchmarks and Metrics.}
We evaluate FocusEarth on four UHR RS-VQA benchmarks.
MME-RealWorld-RS~\cite{zhang2025mmerealworld} contains
3,738 multiple-choice questions covering color,
counting, and spatial position.
LRS-VQA~\cite{luo2025largersi} comprises 7,333 open-ended
questions across eight question categories and three
image-source subsets: FAIR, Bridge, and STAR.
XLRS-Bench-Lite~\cite{wang2025xlrsbench} contains 3,080
questions spanning four perception and four reasoning
task groups.
LRS-GRO~\cite{liu2026zoomearth} comprises 9,734 questions
at the global, region, and object levels.

We report answer accuracy, with overall accuracy
computed as the sample-size-weighted average unless
otherwise specified.
For MME-RealWorld-RS, accuracy is determined by
option-letter matching.
For LRS-VQA, we normalize answer text and apply exact
matching, integer matching, bidirectional substring
matching with a minimum string length of three
characters, or token-multiset $F_1$ matching with a
threshold of 0.8.
For LRS-GRO, we use exact matching or
WordNet~\cite{miller1995wordnet} path similarity after
noun lemmatization, with a similarity threshold of 0.8.
For XLRS-Bench-Lite, single-answer questions require
the correct option, while multiple-answer questions
require exact option-set matching.
Invalid predictions are counted as incorrect.

On XLRS-Bench-Lite, visual representations are
constructed directly from the provided regions for
1,657 questions, while the remaining 1,423 questions
use question-guided region selection.
The 60 temporal questions are evaluated using
chronologically ordered image pairs.
For region localization, we additionally evaluate
the 7,394 LRS-GRO questions with valid target boxes.

\noindent\textbf{Baselines and Implementation Details.}
We compare FocusEarth with general-purpose VLMs,
remote sensing VLMs, and specialized UHR RS-VQA
methods, including both training-based and
training-free approaches.
Unless otherwise specified, we use
Qwen3-VL-8B-Instruct~\cite{bai2025qwen3vl}
as the answer model.
Region scoring employs four ViT-B/32 image--text
encoders: CLIP~\cite{radford2021clip},
GeoRSCLIP~\cite{zhang2024rs5m},
RemoteCLIP~\cite{liu2024remoteclip}, and
SkyCLIP-50~\cite{wang2024skyscript}.
All models are kept frozen throughout inference.

We partition the source image into a $6\times6$
candidate grid with a 5\% per-side expansion.
Each candidate is evaluated using its full-region
view and four corner crops.
The default region budget is set to $K=6$.
The answer model uses greedy decoding with a
maximum of 64 generated tokens.
The compact visual representation is paired with
the original question and answer choices, where
applicable.
\figref{fig:answer_prompt} illustrates the answering
prompt used for MME-RealWorld-RS.

\begin{figure}[!t]
\centering
\begin{FEpromptbox}
\small\rmfamily\raggedright
\textbf{You are an intelligent remote sensing analyst.}\par\smallskip
Image 1 is one source-aligned full-detail image. The entire scene is a direct bilinear resize from the full-resolution source.\par\smallskip
Outlined regions mark selected evidence locations at their original positions. Outlines identify locations only and do not encode the answer.\par\medskip
\textbf{Question:} \texttt{\{question\}}\par\smallskip
Use the following response format:\par\smallskip
\textbf{Evidence:} [A concise, image-grounded observation-based explanation]\par\smallskip
\textbf{Final answer:} [A single MME option letter]
\end{FEpromptbox}
\caption{Prompt template used for single-image answering on MME-RealWorld-RS, specifying the resized full-scene input, the role of region outlines, and the required output format.}
\label{fig:answer_prompt}
\end{figure}

\subsection{Main Results}
\label{sec:main_results}

\begin{table*}[!t]
\centering
\caption{Answer accuracy (\%) on LRS-VQA and MME-RealWorld-RS. LRS-VQA is grouped by source image set, and MME-RealWorld-RS is grouped by question category. Overall accuracy (Avg.) is micro-averaged across all questions. Training-free status is indicated for UHR methods only, and dashes denote unreported results.}
\label{tab:mme_results}
\footnotesize
\setlength{\tabcolsep}{2.5pt}
\renewcommand{\arraystretch}{1.10}
\begin{tabularx}{\textwidth}{@{}>{\raggedright\arraybackslash}p{0.24\textwidth}!{\color{black!35}\vrule width 0.35pt}cc!{\color{black!35}\vrule width 0.35pt}*{3}{>{\centering\arraybackslash}X}!{\color{black!35}\vrule width 0.35pt}>{\centering\arraybackslash}X!{\color{black!35}\vrule width 0.35pt}*{3}{>{\centering\arraybackslash}X}!{\color{black!35}\vrule width 0.35pt}>{\centering\arraybackslash}X@{}}
\toprule
\multicolumn{1}{c!{\color{black!35}\vrule width 0.35pt}}{\multirow{2}{*}{\textbf{Method}}}&\multirow{2}{*}{\textbf{\#Params}}&\multirow{2}{*}{\textbf{Training-free}}&\multicolumn{4}{c!{\color{black!35}\vrule width 0.35pt}}{\textbf{LRS-VQA}}&\multicolumn{4}{c}{\textbf{MME-RealWorld-RS}}\\
\cmidrule(lr){4-7}\cmidrule(l){8-11}
& & &\textbf{FAIR}&\textbf{Bridge}&\textbf{STAR}&\textbf{Avg.}&\textbf{Color}&\textbf{Count}&\textbf{Position}&\textbf{Avg.}\\
\midrule
\rowcolor{red!6}\multicolumn{11}{c}{\textbf{\textit{General-Purpose VLMs}}}\\
Qwen2.5-VL~\cite{bai2025qwen25vl}&7B&\FENa&22.85&36.94&25.78&26.49&45.94&21.05&55.23&40.90\\
Qwen3-VL~\cite{bai2025qwen3vl}&8B&\FENa&24.61&\underline{38.15}&27.53&28.16&48.07&21.35&56.41&42.11\\
LLaVA-OneVision~\cite{li2024llavaov}&7B&\FENa&20.61&35.11&26.08&25.69&26.81&26.14&27.57&26.85\\
LLaVA-v1.5~\cite{liu2024llava15}&7B&\FENa&18.76&30.70&22.63&22.60&23.11&16.88&26.25&22.12\\
LLaVA-v1.6~\cite{liu2024llavanext}&7B&\FENa&19.55&31.97&23.58&23.55&24.51&18.03&27.49&23.39\\
IXC-2.5~\cite{zhang2024ixc25}&7B&\FENa&25.25&\textbf{38.41}&27.30&28.27&45.34&17.62&44.95&36.12\\
\midrule
\rowcolor{green!4}\multicolumn{11}{c}{\textbf{\textit{Remote-Sensing VLMs}}}\\
GeoChat~\cite{kuckreja2024geochat}&7B&\FENa&20.18&24.54&13.75&17.30&23.11&15.66&25.06&21.32\\
VHM~\cite{pang2025vhm}&7B&\FENa&22.59&34.40&26.75&26.57&20.32&16.80&35.24&24.18\\
RSUniVLM~\cite{liu2024rsunivlm}&0.5B&\FENa&21.02&32.61&24.72&24.72&28.97&19.00&28.33&25.48\\
\midrule
\rowcolor{cyan!5}\multicolumn{11}{c}{\textbf{\textit{UHR RS-VQA Methods}}}\\
GeoLLaVA-8K~\cite{wang2025geollava8k}&7B&$\xmark$&20.81&32.73&24.56&24.58&27.92&22.27&34.90&28.41\\
Coarse-to-Fine Pruning~\cite{luo2025largersi}&7B&$\xmark$&22.97&36.89&27.48&27.45&\multicolumn{1}{c}{\FENa}&\multicolumn{1}{c}{\FENa}&\multicolumn{1}{c}{\FENa}&39.04\\
Beyond Zooming~\cite{wang2026beyondzooming}&7B&$\xmark$&\multicolumn{1}{c}{\FENa}&\multicolumn{1}{c}{\FENa}&\multicolumn{1}{c}{\FENa}&\multicolumn{1}{!{\color{black!35}\vrule width 0.35pt}c!{\color{black!35}\vrule width 0.35pt}}{\FENa}&53.23&19.74&63.64&45.75\\
ZoomEarth~\cite{liu2026zoomearth}&3B&$\xmark$&\multicolumn{1}{c}{\FENa}&\multicolumn{1}{c}{\FENa}&\multicolumn{1}{c}{\FENa}&\multicolumn{1}{!{\color{black!35}\vrule width 0.35pt}c!{\color{black!35}\vrule width 0.35pt}}{\FENa}&49.69&\textbf{31.66}&50.14&43.93\\
ZoomEye~\cite{shen2025zoomeye}&7B&$\cmark$&22.53&34.48&26.27&26.30&48.54&17.76&31.18&32.61\\
ZoomSearch~\cite{zhou2025zoomsearch}&7B&$\cmark$&21.83&32.33&26.05&25.65&48.62&20.05&59.83&43.02\\
WeaveEarth~\cite{ma2026weaveearth}&8B&$\cmark$&\underline{30.99}&26.08&\underline{36.68}&\underline{33.38}&51.39&\underline{27.65}&62.60&47.38\\
GazeEarth~\cite{zhang2026gazeearth}&8B&$\cmark$&\multicolumn{1}{c}{\FENa}&\multicolumn{1}{c}{\FENa}&\multicolumn{1}{c}{\FENa}&\multicolumn{1}{!{\color{black!35}\vrule width 0.35pt}c!{\color{black!35}\vrule width 0.35pt}}{\FENa}&\underline{63.67}&21.78&\underline{71.12}&\underline{52.43}\\
\midrule
\rowcolor{black!6}\textbf{FocusEarth (ours)}&8B&$\cmark$&\textbf{31.12}&32.02&\textbf{36.76}&\textbf{34.32}&\textbf{65.34}&26.35&\textbf{75.34}&\textbf{55.91}\\
\bottomrule
\end{tabularx}
\end{table*}

\begin{figure*}[!t]
\centering
\textbf{(a) Spatial-position question}\par\vspace{1mm}
\includegraphics[width=0.92\textwidth]{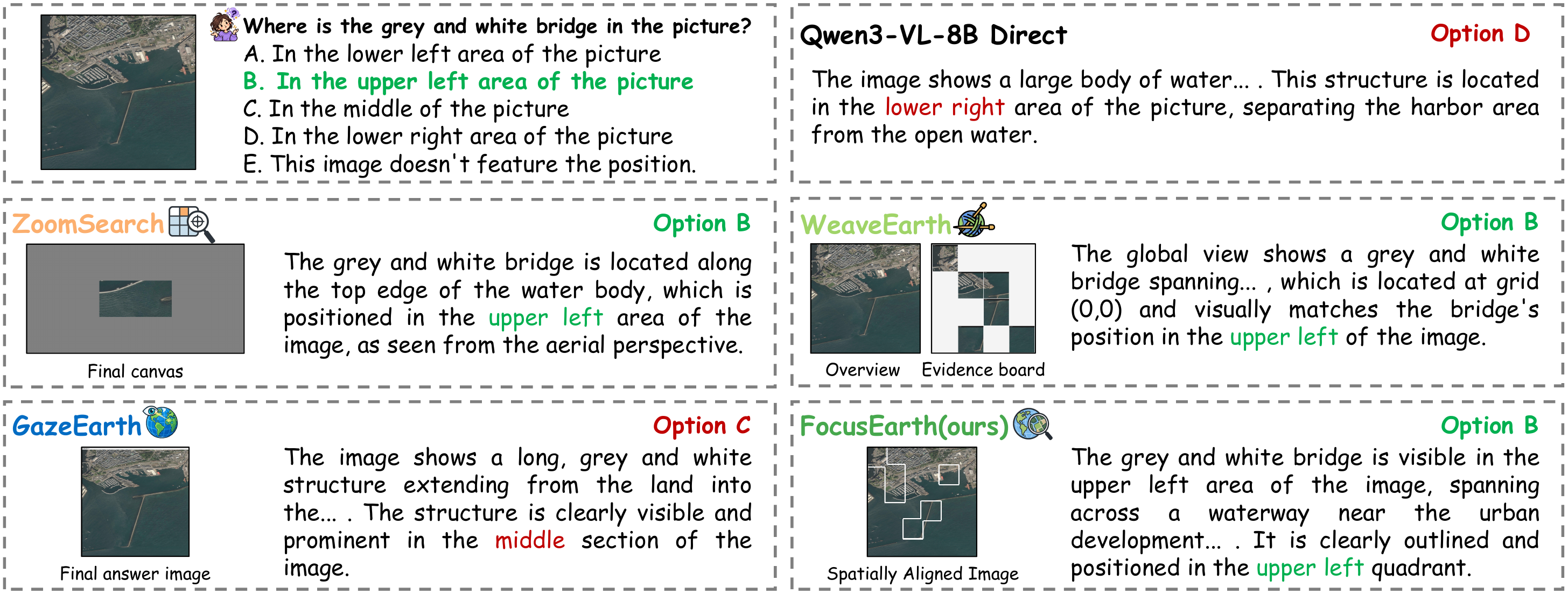}\par\vspace{2mm}
\textbf{(b) Color question}\par\vspace{1mm}
\includegraphics[width=0.92\textwidth]{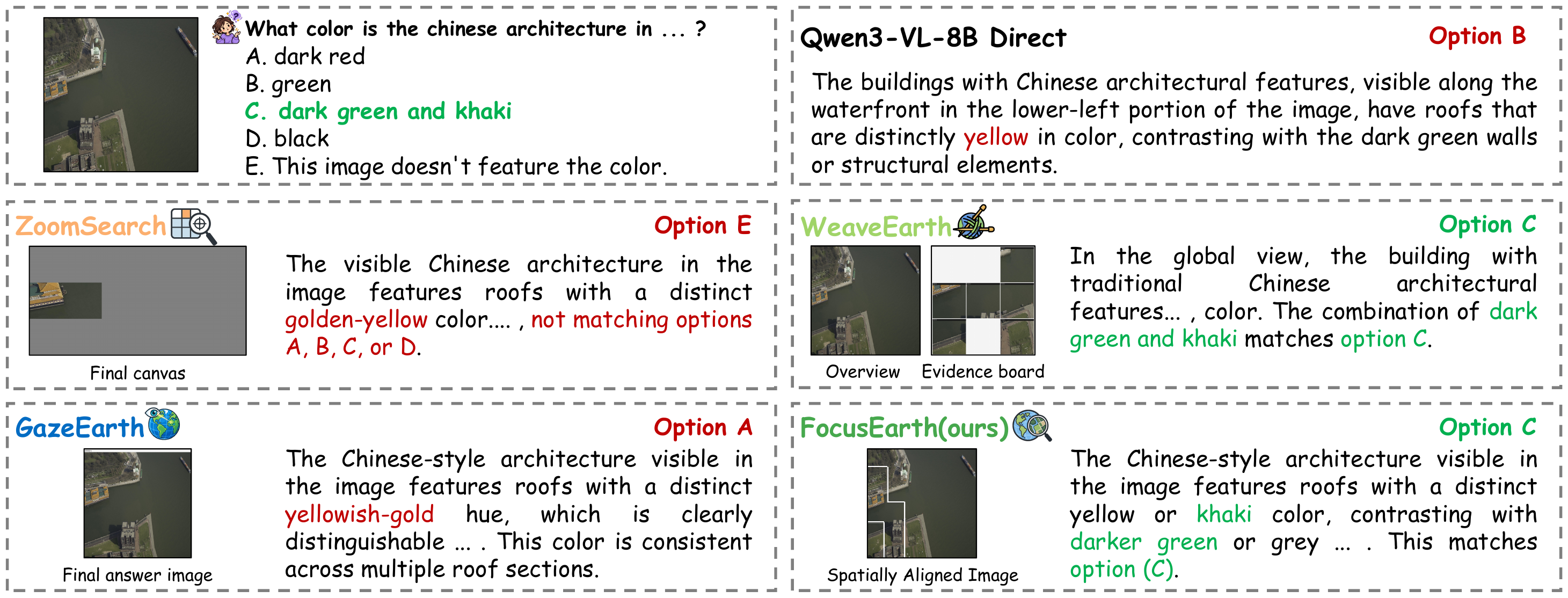}
\caption{Qualitative comparison of visual representations and answer predictions on MME-RealWorld-RS using the same frozen Qwen3-VL-8B backbone. (a) Spatial-position question. (b) Color question.}
\label{fig:qualitative_examples}
\end{figure*}

\begin{table*}[!t]
\centering
\caption{Answer accuracy (\%) on XLRS-Bench-Lite across four perception and four reasoning task groups. Avg.\ denotes micro-averaged accuracy across all 3,080 questions. Cnt: counting, SC: scene classification, OSR: object spatial relationship, OP: object properties, Plan: planning, AR: anomaly reasoning, CR: complex reasoning, SR: spatiotemporal reasoning.}
\label{tab:xlrs}
\footnotesize
\setlength{\tabcolsep}{3pt}
\renewcommand{\arraystretch}{1.10}
\begin{tabularx}{\textwidth}{ @{} >{\raggedright\arraybackslash}p{0.27\textwidth} !{\color{black!35}\vrule width 0.35pt} *{4}{>{\centering\arraybackslash}X} !{\color{black!35}\vrule width 0.35pt} *{4}{>{\centering\arraybackslash}X} !{\color{black!35}\vrule width 0.35pt} >{\centering\arraybackslash}X @{} }
\toprule
\multicolumn{1}{c!{\color{black!35}\vrule width 0.35pt}} {\multirow{2}{*}{\textbf{Method}}} & \multicolumn{4}{c!{\color{black!35}\vrule width 0.35pt}} {\textbf{Perception}} & \multicolumn{4}{c!{\color{black!35}\vrule width 0.35pt}} {\textbf{Reasoning}} & \multirow{2}{*}{\textbf{Avg.}}\\
\cmidrule(lr){2-5}\cmidrule(lr){6-9}
&\textbf{Cnt}&\textbf{SC}&\textbf{OSR}&\textbf{OP}&\textbf{Plan}&\textbf{AR}&\textbf{CR}&\textbf{SR}&\\
\midrule
\rowcolor{red!6}\multicolumn{10}{c}{\textbf{\textit{General-Purpose VLMs}}}\\
Qwen2.5-VL-7B~\cite{bai2025qwen25vl}&30.63&40.33&30.80&43.31&32.00&67.00&62.50&43.33&41.98\\
Qwen3-VL-8B~\cite{bai2025qwen3vl}&32.50&41.67&31.40&43.92&34.00&70.00&63.50&46.33&42.92\\
LLaVA-OneVision-7B~\cite{li2024llavaov}&31.25&35.67&25.20&45.42&24.00&76.00&59.50&43.33&41.62\\
InternVL3-8B~\cite{zhu2025internvl3}&36.88&40.67&25.20&46.80&36.00&\textbf{77.00}&66.00&21.67&43.57\\
\midrule
\rowcolor{green!4}\multicolumn{10}{c}{\textbf{\textit{Remote-Sensing VLMs}}}\\
GeoChat~\cite{kuckreja2024geochat}&22.50&12.67&24.20&24.28&10.00&33.00&32.00&\multicolumn{1}{c}{\FENa}&23.35\\
\midrule
\rowcolor{cyan!5}\multicolumn{10}{c}{\textbf{\textit{UHR RS-VQA Methods}}}\\
GeoLLaVA-8K~\cite{wang2025geollava8k}&35.00&62.00&35.80&37.71&\textbf{65.00}&68.00&63.50&51.67&43.44\\
Beyond Zooming~\cite{wang2026beyondzooming}&36.25&62.33&34.40&45.96&58.00&66.00&65.00&48.33&47.50\\
WeaveEarth~\cite{ma2026weaveearth}&37.50&50.33&36.40&47.41&36.00&73.00&67.50&46.33&47.14\\
GazeEarth~\cite{zhang2026gazeearth}&\textbf{44.38}&66.67&\textbf{37.80}&53.07&49.00&76.00&\textbf{74.50}&\textbf{55.00}&\textbf{53.51}\\
\midrule
\rowcolor{black!6}\textbf{FocusEarth (ours)}&36.88&\textbf{68.00}&29.60&\textbf{57.05}&36.00&74.00&67.50&48.33&52.99\\
\bottomrule
\end{tabularx}
\end{table*}

\begin{table}[t]
\centering
\caption{Answer accuracy (\%) on LRS-GRO across global, region, and object levels. Avg.\ denotes micro-averaged accuracy across all questions.}
\label{tab:lrs_gro}
\footnotesize
\setlength{\tabcolsep}{3pt}
\renewcommand{\arraystretch}{1.10}
\begin{tabularx}{\columnwidth}{@{}>{\raggedright\arraybackslash}X!{\color{black!35}\vrule width 0.35pt}*{3}{c}!{\color{black!35}\vrule width 0.35pt}c@{}}
\toprule
\multicolumn{1}{c!{\color{black!35}\vrule width 0.35pt}}{\textbf{Method}}&\textbf{Global}&\textbf{Region}&\textbf{Object}&\textbf{Avg.}\\
\midrule
\rowcolor{red!6}\multicolumn{5}{c}{\textbf{\textit{General-Purpose VLMs}}}\\
IXC-2.5~\cite{zhang2024ixc25}&62.43&40.00&47.32&50.00\\
Qwen2.5-VL-3B~\cite{bai2025qwen25vl}&58.90&31.76&38.66&42.83\\
Qwen2.5-VL-7B~\cite{bai2025qwen25vl}&70.06&39.08&46.43&51.43\\
InternVL3-8B~\cite{zhu2025internvl3}&71.60&44.58&47.80&53.67\\
VLM-$R^{3}$~\cite{jiang2025vlmr3} (tools)&69.72&44.83&37.40&50.17\\
\midrule
\rowcolor{green!4}\multicolumn{5}{c}{\textbf{\textit{Remote-Sensing VLMs}}}\\
GeoChat~\cite{kuckreja2024geochat}&62.43&36.79&40.72&45.88\\
\midrule
\rowcolor{cyan!5}\multicolumn{5}{c}{\textbf{\textit{UHR RS-VQA Methods}}}\\
Beyond Zooming~\cite{wang2026beyondzooming}&\textbf{79.00}&48.90&57.60&60.67\\
ZoomEarth~\cite{liu2026zoomearth}&63.09&46.11&51.80&53.76\\
GazeEarth~\cite{zhang2026gazeearth}&77.96&\underline{50.39}&\underline{58.04}&\underline{61.00}\\
\midrule
\rowcolor{black!6}\textbf{FocusEarth (ours)}&\underline{78.57}&\textbf{51.87}&\textbf{60.63}&\textbf{62.85}\\
\bottomrule
\end{tabularx}
\end{table}

\noindent\textbf{MME-RealWorld-RS.}
As shown in \tabref{tab:mme_results}, FocusEarth
achieves the highest overall accuracy of 55.91\%
among the compared methods, improving upon direct
Qwen3-VL-8B inference by 13.80 percentage points
without additional model training.
Compared with recent training-free UHR methods,
FocusEarth improves overall accuracy over GazeEarth
and WeaveEarth by 3.48 and 8.53 percentage points,
respectively.
The largest gains over direct inference are observed
on Color and Position, where accuracy reaches
65.34\% and 75.34\%, respectively, corresponding to
improvements of 17.27 and 18.93 percentage points.
The gains in these two categories are consistent
with our design of retaining local visual features
within the full-scene spatial layout.
Counting remains more challenging: although
FocusEarth improves accuracy by 5.00 percentage
points over direct inference, its accuracy of
26.35\% remains below that of ZoomEarth and
WeaveEarth.
Representative counting failures are illustrated
in \figref{fig:counting_failures}.

\noindent\textbf{LRS-VQA.}
On LRS-VQA (\tabref{tab:mme_results}), FocusEarth
achieves the highest overall accuracy of 34.32\%,
exceeding direct Qwen3-VL-8B inference and
WeaveEarth by 6.16 and 0.94 percentage points,
respectively.
FocusEarth achieves the highest accuracy on FAIR
and STAR, with gains of 6.51 and 9.23 percentage
points over direct inference, respectively.
On Bridge, it improves upon WeaveEarth by 5.94
percentage points but remains below direct
Qwen3-VL-8B inference.
The category-level results in
\tabref{tab:lrs_categories} show that FocusEarth
achieves 61.90\% accuracy on rural--urban
classification, while attaining 49.02\% and
36.05\% on object color and shape recognition,
respectively.
Consistent with the results on MME-RealWorld-RS,
counting remains the most challenging category,
with an accuracy of 16.01\% over 1,199 questions.
The macro-average across the eight question
categories is 33.47\%.

\noindent\textbf{XLRS-Bench-Lite.}
FocusEarth attains an overall accuracy of 52.99\%
on XLRS-Bench-Lite (\tabref{tab:xlrs}), exceeding
Qwen3-VL-8B and WeaveEarth by 10.07 and 5.85
percentage points, respectively.
Its overall accuracy is 0.52 percentage points
below that of GazeEarth.
At the task level, FocusEarth leads on Scene
Classification and Object Properties, with
accuracies of 68.00\% and 57.05\%, respectively,
but achieves lower accuracies than GazeEarth
on Counting, Object Spatial Relationship, and
the four reasoning task groups.
For questions with provided regions, FocusEarth
obtains 57.15\% accuracy (947/1,657), compared
with 48.14\% (685/1,423) on questions requiring
automatic region selection.

\noindent\textbf{LRS-GRO.}
With an overall accuracy of 62.85\%, FocusEarth
ranks first among the methods compared on LRS-GRO
(\tabref{tab:lrs_gro}), ahead of GazeEarth and
Beyond Zooming by 1.85 and 2.18 percentage points,
respectively.
Its advantage over GazeEarth is more pronounced at
finer spatial scales.
FocusEarth obtains the highest accuracies of
51.87\% and 60.63\% on Region and Object questions,
respectively, exceeding GazeEarth by 1.48 and 2.59
percentage points.
At the global level, FocusEarth obtains 78.57\%
accuracy, compared with 79.00\% for Beyond Zooming
and 77.96\% for GazeEarth.
The overall advantage is primarily associated with
the stronger performance on region- and object-level
questions.

\noindent\textbf{Qualitative Examples.}
\figref{fig:qualitative_examples} compares different
visual representations using the same frozen
Qwen3-VL-8B answer model.
Except for direct full-image inference, all
representations are constructed from the same
selected regions.
In \figref{fig:qualitative_examples}(a), direct
inference places the gray-and-white bridge in the
lower-right area, whereas the GazeEarth-style
foveated representation leads to a prediction
in the middle of the image.
FocusEarth preserves the bridge within its original
scene layout and correctly identifies its upper-left
location, as do the ZoomSearch-style and
WeaveEarth-style representations.
In \figref{fig:qualitative_examples}(b), the color
question requires identifying both dark green
and khaki in the Chinese architecture.
Direct inference and the GazeEarth-style
representation return incomplete color
combinations, while the ZoomSearch-style
representation selects the negative option.
Both FocusEarth and the WeaveEarth-style
representation identify the correct color
combination.
The two examples illustrate how FocusEarth
retains question-relevant local features while
preserving the spatial layout of the full scene.

\begin{table}[!t]
\centering
\caption{\normalfont LRS-VQA accuracy by question category (\%).}
\label{tab:lrs_categories}
\footnotesize
\begin{tabularx}{\columnwidth}{@{}>{\raggedright\arraybackslash}Xrr@{}}
\toprule
\textbf{Category} &
\textbf{Correct / Total} &
\textbf{Acc. (\%)} \\
\midrule
Count                      & 192/1,199 & 16.01 \\
Object background          & 61/245    & 24.90 \\
Object category            & 281/987   & 28.47 \\
Object color               & 375/765   & 49.02 \\
Object shape               & 319/885   & 36.05 \\
Object status              & 232/1,000 & 23.20 \\
Reasoning                  & 282/1,000 & 28.20 \\
Rural or urban             & 775/1,252 & 61.90 \\
\midrule
Question-weighted overall  & 2,517/7,333 & 34.32 \\
Category macro-average     & -- & 33.47 \\
\bottomrule
\vspace{-0.3cm}
\end{tabularx}
\end{table}

\subsection{Ablation Study}
\label{sec:ablations}

\subsubsection{Component Controls}

\tabref{tab:region_ablation} presents component
ablation results across the three stages of
FocusEarth on MME-RealWorld-RS.
In Stage 1, replacing the reformulated context
description with the original question text
without answer choices
(\textit{w/o Context Reformulation}) reduces
overall accuracy by 1.12 percentage points.
Replacing the four scoring encoders and
entropy-matched fusion with a single SigLIP 2
encoder~\cite{tschannen2025siglip2}
(\textit{w/o Encoder Ensemble}) results in a
reduction of 3.45 percentage points.
In Stage 2, selecting regions solely according
to the context ranking without entity proposals
(\textit{w/o Entity Proposals}) lowers accuracy
by 0.83 percentage points under the same region
budget.
All three variants exhibit lower accuracy across
Color, Count, and Position.

Among the Stage 3 variants, replacing the
spatially aligned full-scene image with a
thumbnail-and-patch montage
(\textit{w/o Spatial Alignment}) results in the
largest overall accuracy drop of 6.71 percentage
points.
Position accuracy decreases from 75.34\% to
55.53\%, whereas Count improves from 26.35\%
to 28.38\%, highlighting the importance of
preserving scene-level spatial references for
position-related questions.
Removing token protection
(\textit{w/o Token Protection}) reduces overall
accuracy by 1.18 percentage points, while removing
region outlines and their corresponding prompt
cues (\textit{w/o Region Outlines}) results in a
reduction of 1.23 percentage points.
Removing both lowers accuracy to 53.29\%,
2.62 percentage points below the full method.

The effect of question-guided region selection
is further examined against Random-6, which
randomly selects six of the 36 candidate regions
with a fixed seed while retaining the same
visual representation construction and token
compression procedures.
FocusEarth achieves 55.91\% accuracy
(2,090/3,738), compared with 52.49\%
(1,962/3,738) for Random-6.
Across paired predictions, FocusEarth correctly
answers 291 questions missed by Random-6,
whereas the reverse holds for 163 questions.
The resulting accuracy difference of 3.42
percentage points has a 95\% confidence interval
of [2.32, 4.53] percentage points when clustered
by source image.

To examine the contribution of multi-encoder
fusion, we evaluate each constituent encoder
individually under identical candidate regions,
scoring views, query descriptions, selection
rules, and answer model settings.
GeoRSCLIP achieves the highest single-encoder
accuracy of 55.35\%, followed by SkyCLIP-50
(55.24\%), CLIP (54.92\%), and RemoteCLIP
(53.77\%).
The fused scores yield an overall accuracy of
55.91\%, exceeding the strongest individual
encoder by 0.56 percentage points, with consistent
gains across Color, Count, and Position.

\begin{table}[!t]
\centering
\caption{Component ablation across the three stages of FocusEarth on MME-RealWorld-RS (accuracy, \%). $\Delta$ denotes percentage-point change relative to FocusEarth (full).}
\label{tab:region_ablation}
\footnotesize
\setlength{\tabcolsep}{2.5pt}
\renewcommand{\arraystretch}{1.10}
\begin{tabularx}{\columnwidth}{@{}>{\raggedright\arraybackslash}Xccccc@{}}
\toprule
\multicolumn{1}{c}{\textbf{Variant}}&\textbf{Color}&\textbf{Count}&\textbf{Position}&\textbf{Avg.}&\textbf{$\Delta$ (pp)}\\
\midrule
\rowcolor{black!4}\multicolumn{6}{@{}l}{\textbf{Stage 1: Query-Adaptive Region Prioritization}}\\
w/o Context Reformulation&63.82&25.61&74.22&54.79&$-1.12$\\
w/o Encoder Ensemble&62.31&22.43&71.92&52.46&$-3.45$\\
\rowcolor{black!4}\multicolumn{6}{@{}l}{\textbf{Stage 2: Multi-Query Region Selection}}\\
w/o Entity Proposals&64.30&25.86&74.38&55.08&$-0.83$\\
\rowcolor{black!4}\multicolumn{6}{@{}l}{\textbf{Stage 3: Spatially Aligned Visual Representation}}\\
w/o Spatial Alignment&63.19&\textbf{28.38}&55.53&49.20&$-6.71$\\
w/o Token Protection&64.22&24.63&74.62&54.74&$-1.18$\\
w/o Region Outlines&63.98&25.86&73.51&54.68&$-1.23$\\
w/o Outlines and Protection&63.03&24.23&71.92&53.29&$-2.62$\\
\midrule
\rowcolor{black!6}\textbf{FocusEarth (full)}&\textbf{65.34}&26.35&\textbf{75.34}&\textbf{55.91}&--\\
\bottomrule
\end{tabularx}
\vspace{-2mm}
\end{table}

\subsubsection{Feature Protection Alignment}

To examine the effect of feature protection
at matched token budgets,
\tabref{tab:mask_seed_ablation} compares variants
with identical selected regions, encoded visual
features, answer prompts, and per-question
visual-token counts.
Aligned protection preserves feature cells
corresponding to the selected regions, whereas
misaligned protection uses a transformed mask
with the same number of protected cells.
The misaligned mask is obtained from the flip
or rotation that minimizes overlap with the
selected regions.
Both protection conditions are evaluated with
Morton-order and row-major seeding.
Native Uniform aggregates visual features
without region-specific protection under the
same token budget.

Aligned protection improves overall accuracy
over misaligned protection by 1.47 and 0.83
percentage points under Morton-order and
row-major seeding, respectively.
These results highlight the importance of
spatial correspondence between selected regions
and protected visual features.
Under aligned protection, Morton-order seeding
further improves accuracy by 0.48 percentage
points over row-major seeding, although this
advantage is not observed under misaligned
protection.
Compared with Native Uniform, FocusEarth
improves overall accuracy from 55.00\% to
55.91\%, with the largest gain observed on
Count.

The outline and protection variants in
\tabref{tab:region_ablation} provide a
complementary comparison.
Relative to removing both components, retaining
only region outlines with their prompt cues
improves accuracy by 1.45 percentage points,
while retaining only token protection yields
a gain of 1.39 percentage points.
Combining both achieves the highest accuracy
of 55.91\%.

\begin{table}[!t]
\centering
\setlength{\abovecaptionskip}{1.5pt}
\caption{Effect of protection-mask alignment and grouping seed order on MME-RealWorld-RS (accuracy, \%) under matched per-question visual token budgets. $\Delta$ is relative to FocusEarth (full).}
\label{tab:mask_seed_ablation}
\footnotesize
\setlength{\tabcolsep}{1.6pt}
\renewcommand{\arraystretch}{1.14}
\setlength{\aboverulesep}{1.4pt}
\setlength{\belowrulesep}{1.6pt}
\newcommand{\maskseedhead}[1]{\ensuremath{\vcenter{\hbox{\bfseries\shortstack[c]{#1}}}}}
\begin{tabularx}{\columnwidth}{
  >{\raggedright\arraybackslash}X
  c c
  c c c >{\hspace{1.4pt}}c
  >{\hspace{1.0pt}}c
}
\toprule
\maskseedhead{Variant} &
\maskseedhead{Aligned\\mask} &
\maskseedhead{Morton\\seeds} &
\maskseedhead{Color} &
\maskseedhead{Count} &
\maskseedhead{Position} &
\maskseedhead{Avg.} &
\maskseedhead{$\boldsymbol{\Delta}$ (pp)} \\
\midrule
Native Uniform   & --           & --           & 64.62 & 24.14 & \textbf{75.50} & 55.00 & $-0.91$ \\
Both changes     & $\times$     & $\times$     & 64.62 & 24.31 & 74.14 & 54.60 & $-1.31$ \\
Misaligned mask  & $\times$     & $\checkmark$ & 64.70 & 24.14 & 73.75 & 54.44 & $-1.47$ \\
Row-major seeds  & $\checkmark$ & $\times$     & 65.18 & 25.53 & 74.86 & 55.43 & $-0.48$ \\
\midrule
\rowcolor{black!6}
\textbf{FocusEarth (full)} &
$\checkmark$ & $\checkmark$ &
\textbf{65.34} & \textbf{26.35} & 75.34 & \textbf{55.91} & -- \\
\bottomrule
\end{tabularx}
\end{table}

\begin{figure}[!t]
\centering
\includegraphics[width=\columnwidth]{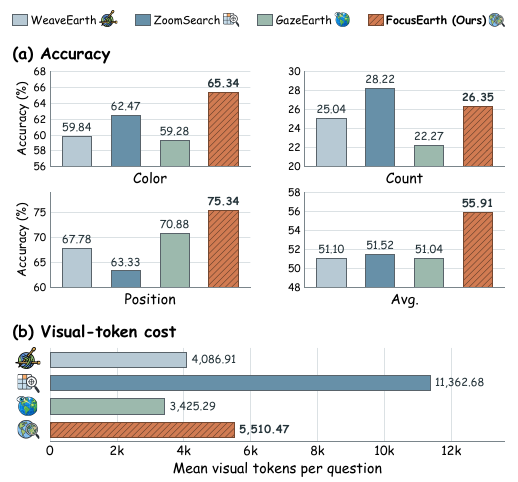}
\caption{Comparison of visual representations on
MME-RealWorld-RS with fixed selected regions and
answer model.
(a) Category-level and overall accuracy.
(b) Mean visual-token count per question.
The alternatives use adapted visual representations
and corresponding prompts of existing methods.}
\label{fig:representation}
\vspace{-2mm}
\end{figure}

With the selected source regions and answer model
fixed, \figref{fig:representation} compares different
visual representations in terms of answer accuracy
and visual-token cost.
FocusEarth achieves the highest overall accuracy
of 55.91\%, leading on both Color (65.34\%) and
Position (75.34\%).
The ZoomSearch-style representation achieves the
highest Count accuracy of 28.22\%, but its Position
accuracy is 12.01 percentage points below that of
FocusEarth.
In terms of visual-token cost, FocusEarth uses
51.5\% fewer tokens than the ZoomSearch-style
representation while attaining an overall accuracy
4.39 percentage points higher.
The WeaveEarth-style and GazeEarth-inspired
representations use fewer visual tokens but yield
lower overall accuracy.

\begin{table}[!t]
\centering
\caption{Region localization accuracy on the 7,394 LRS-GRO questions with valid target boxes. Center Hit@1 and Center Hit@6 measure whether the target center falls inside the top-1 region or the union of the six selected regions, respectively, and Box Coverage@6 reports the mean fraction of the target box covered by the selected set.}
\label{tab:lrsgro_localization}
\footnotesize
\setlength{\tabcolsep}{2.5pt}
\renewcommand{\arraystretch}{1.10}
\begin{tabularx}{\columnwidth}{@{}>{\raggedright\arraybackslash}Xccc@{}}
\toprule
\multirow{2}{*}{\textbf{Variant}}
& \textbf{Center}
& \textbf{Center}
& \textbf{Box} \\
& \textbf{Hit@1 (\%)} $\uparrow$
& \textbf{Hit@6 (\%)} $\uparrow$
& \textbf{Coverage@6 (\%)} $\uparrow$ \\
\midrule
Random-6 & 3.31 & 19.08 & 18.99 \\
CLIP~\cite{radford2021clip} & 34.47 & 74.68 & 69.30 \\
RemoteCLIP~\cite{liu2024remoteclip} & 29.12 & 70.48 & 65.11 \\
GeoRSCLIP~\cite{zhang2024rs5m} & 41.20 & 80.04 & 74.06 \\
SkyCLIP-50~\cite{wang2024skyscript} & 39.59 & 79.17 & 73.72 \\
\midrule
\rowcolor{black!6}\textbf{FocusEarth (ours)} & \textbf{45.36} & \textbf{84.62} & \textbf{78.14} \\
\bottomrule
\end{tabularx}
\vspace{-2mm}
\end{table}

\subsubsection{Region Localization}

We evaluate region localization independently of
answer generation on 7,394 LRS-GRO questions with
valid target boxes (\tabref{tab:lrsgro_localization}).
FocusEarth achieves a Center Hit@1 of 45.36\%,
which increases to 84.62\% when all six selected
regions are considered.
The selected regions also cover an average of
78.14\% of the annotated target area.
Under the same candidate grid and region budget,
FocusEarth outperforms all four constituent
encoders across the three localization metrics.

\begin{figure*}[!t]
\centering
\textbf{(a) Responses to ``largest highway bridge''}\par\vspace{1mm}
\includegraphics[width=0.9\textwidth]{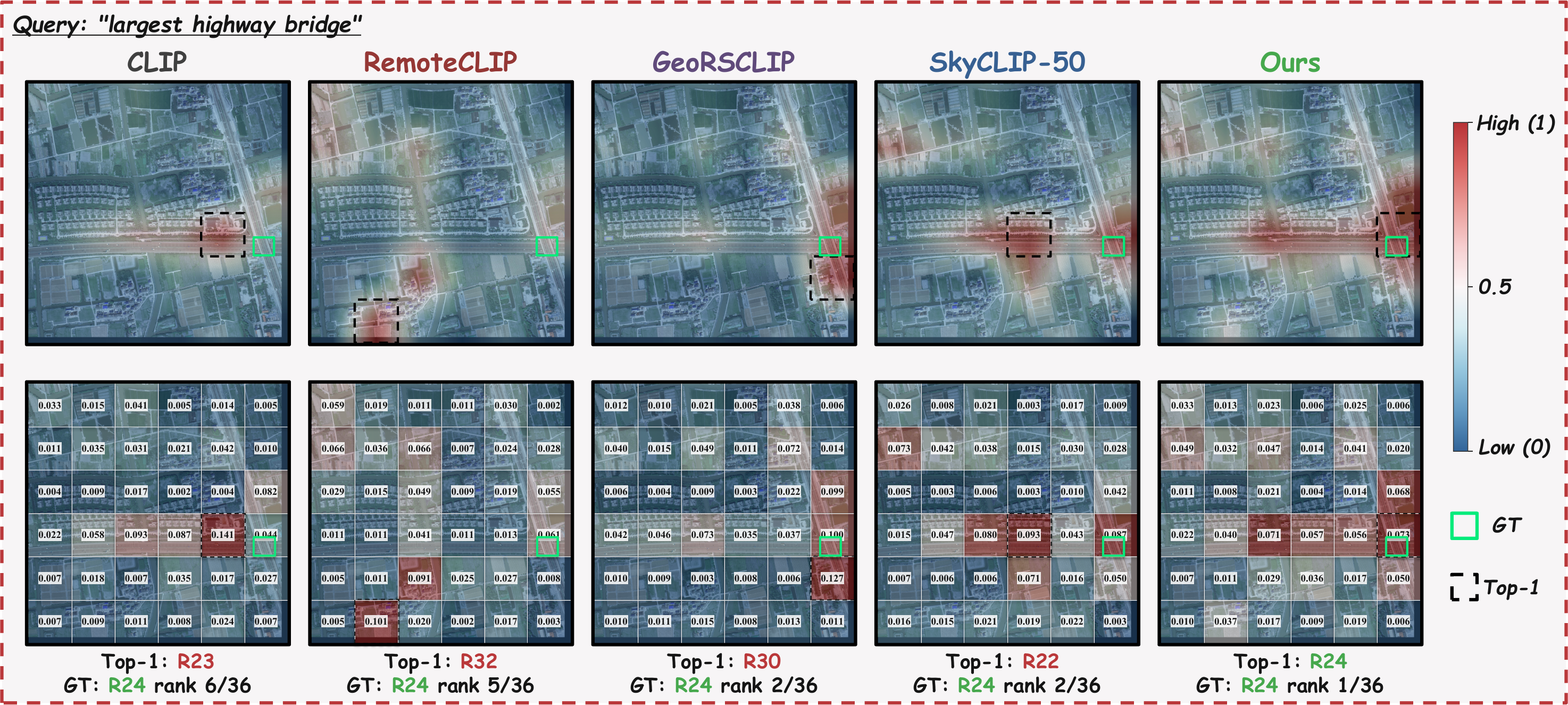}\par\vspace{2mm}
\textbf{(b) Responses to ``top-most water reservoir''}\par\vspace{1mm}
\includegraphics[width=0.9\textwidth]{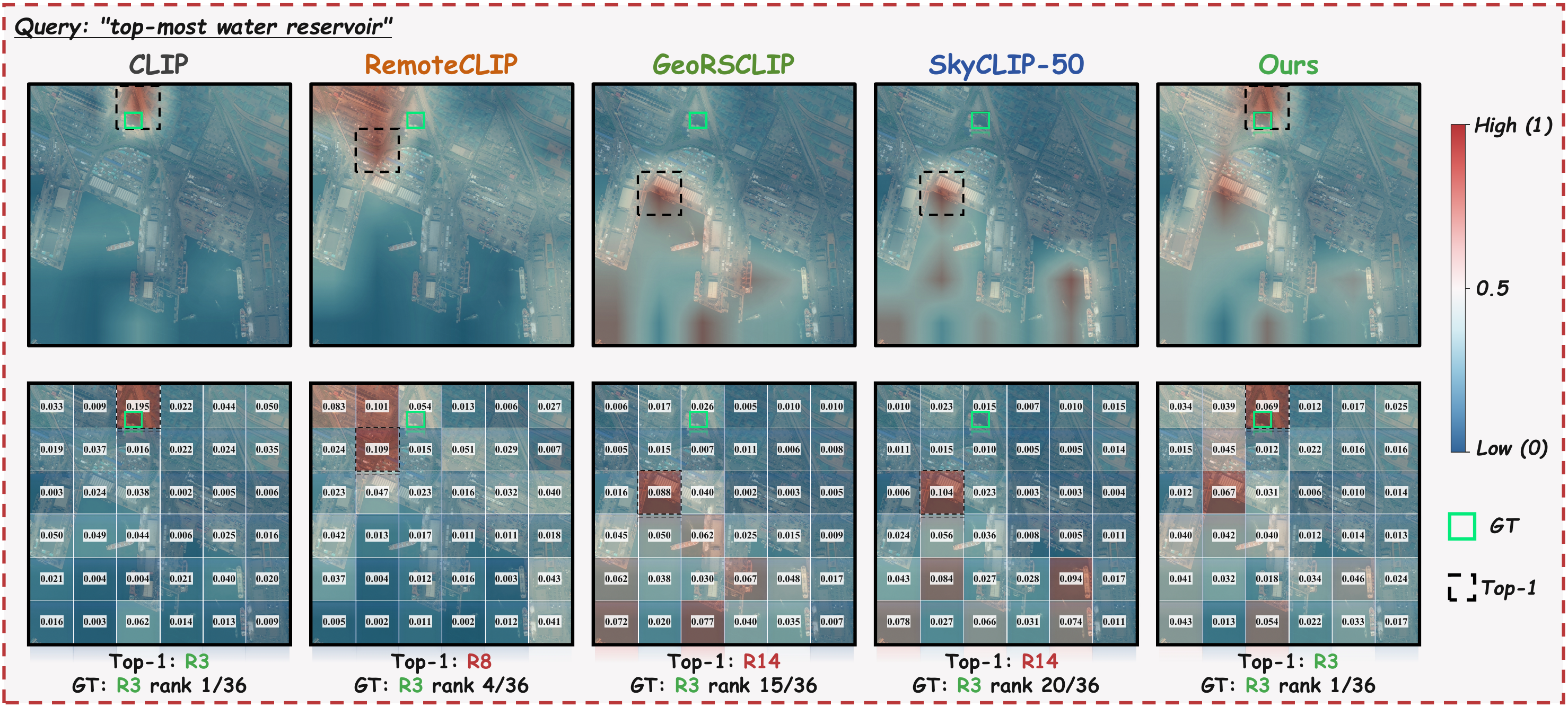}
\caption{Normalized spatial response maps of the four frozen image--text encoders and their entropy-calibrated fusion over the 36 candidate regions for (a) ``largest highway bridge'' and (b) ``top-most water reservoir''. Green boxes mark the ground-truth region, dashed boxes mark the top-1 response, and labels below each map indicate the ground-truth rank.}
\label{fig:region_rankings}
\end{figure*}

\figref{fig:region_rankings} visualizes the regional
responses of individual encoders and their fusion.
For the largest highway bridge in
\figref{fig:region_rankings}(a), the fused
distribution ranks the target region first,
compared with second for GeoRSCLIP and sixth
for CLIP.
For the top-most water reservoir in
\figref{fig:region_rankings}(b), fusion retains
the first-place ranking obtained by CLIP,
whereas RemoteCLIP, GeoRSCLIP, and SkyCLIP-50
rank the target region fourth, fifteenth,
and twentieth, respectively.
The two cases illustrate complementary regional
responses across encoders and the ability of
fusion to improve or retain favorable target
rankings.

\begin{table}[!t]
\centering
\caption{Transfer to frozen answer models on MME-RealWorld-RS (accuracy, \%). Baseline denotes direct full-image inference, and $\Delta$ Acc.\ reports the percentage-point gain for each answer model.}
\label{tab:backbone_transfer_mme}
\footnotesize
\setlength{\tabcolsep}{3pt}
\renewcommand{\arraystretch}{1.07}
\begin{tabularx}{\columnwidth}{@{}>{\raggedright\arraybackslash}Xcccc@{}}
\toprule
\textbf{Method} & \textbf{Color} & \textbf{Count} & \textbf{Position} & \textbf{Avg.} \\
\midrule
\multicolumn{5}{c}{\textbf{\textit{Qwen3-VL-8B}}} \\
Baseline & 48.07 & 21.35 & 56.41 & 42.11 \\
\rowcolor{black!6}\textbf{+FocusEarth} & \textbf{65.34} & \textbf{26.35} & \textbf{75.34} & \textbf{55.91} \\
\textit{$\Delta$ Acc.} & \textcolor{gainpositive}{+17.27} & \textcolor{gainpositive}{+5.00} & \textcolor{gainpositive}{+18.93} & \textcolor{gainpositive}{+13.80} \\
\midrule
\multicolumn{5}{c}{\textbf{\textit{LLaVA-v1.6-Mistral-7B}}} \\
Baseline & 24.51 & \textbf{18.03} & 27.49 & 23.39 \\
\rowcolor{black!6}\textbf{+FocusEarth} & \textbf{39.60} & 14.11 & \textbf{42.72} & \textbf{32.29} \\
\textit{$\Delta$ Acc.} & \textcolor{gainpositive}{+15.09} & \textcolor{gainnegative}{-3.92} & \textcolor{gainpositive}{+15.23} & \textcolor{gainpositive}{+8.90} \\
\midrule
\multicolumn{5}{c}{\textbf{\textit{Intern-S1-mini}}~\cite{bai2025interns1}} \\
Baseline & 45.26 & 13.30 & \textbf{66.11} & 41.79 \\
\rowcolor{black!6}\textbf{+FocusEarth} & \textbf{48.53} & \textbf{23.33} & 65.47 & \textbf{45.96} \\
\textit{$\Delta$ Acc.} & \textcolor{gainpositive}{+3.27} & \textcolor{gainpositive}{+10.03} & \textcolor{gainnegative}{-0.64} & \textcolor{gainpositive}{+4.17} \\
\bottomrule
\end{tabularx}
\vspace{-2mm}
\end{table}

\subsubsection{Transfer to Other Answer Models}

Beyond Qwen3-VL-8B, we instantiate FocusEarth
with two additional frozen answer models:
LLaVA-v1.6-Mistral-7B and Intern-S1-mini.
Region selection remains unchanged, while feature
compression follows the native visual feature
organization of each answer model.
As reported in \tabref{tab:backbone_transfer_mme},
FocusEarth exceeds the corresponding direct-inference
baselines by 13.80, 8.90, and 4.17 percentage points
for Qwen3-VL-8B, LLaVA-v1.6-Mistral-7B, and
Intern-S1-mini, respectively.
Color accuracy is consistently higher across
the three models, whereas the changes on Count
and Position vary with the answer model.
Intern-S1-mini obtains a 10.03-point improvement
on Count but a 0.64-point decrease on Position,
while LLaVA-v1.6-Mistral-7B exhibits a 3.92-point
decrease on Count.
These results support the applicability of
FocusEarth across different frozen answer models.

\subsection{Inference Efficiency}
\label{sec:efficiency}

\begin{table}[!t]
\centering
\caption{Answer accuracy, mean visual-token count,
and per-question answering-stage latency on
MME-RealWorld-RS.
Visual-token counts refer to the sequences delivered
to the language model.
Prep., TTFT, and Ans.\ denote input preparation,
time to first token, and answer generation,
respectively.
Region scoring and selection are precomputed offline.
Accuracy and latency are measured in separate runs.}
\label{tab:cost}
\scriptsize
\setlength{\tabcolsep}{2pt}
\renewcommand{\arraystretch}{1.10}
\begin{tabularx}{\columnwidth}{@{}>{\raggedright\arraybackslash}X*{6}{c}@{}}
\toprule
\multirow{2}{*}{\textbf{Method}}
& \multicolumn{1}{c}{Accuracy}
& \multicolumn{1}{c}{Visual Tokens}
& \multicolumn{4}{c}{Latency (s/question) $\downarrow$} \\
\cmidrule(lr){2-2}\cmidrule(lr){3-3}\cmidrule(lr){4-7}
& \multicolumn{1}{c}{Acc. (\%) $\uparrow$}
& \multicolumn{1}{c}{Count $\downarrow$}
& \multicolumn{1}{c}{Prep.}
& \multicolumn{1}{c}{TTFT}
& \multicolumn{1}{c}{Ans.}
& \multicolumn{1}{c}{Total} \\
\midrule
Direct Qwen2.5-VL-3B & 23.94 & 15,670.51 & 0.531 & 2.376 & 0.018 & 2.925 \\
\rowcolor{black!6}\textbf{+FocusEarth} & \textbf{34.19} & \textbf{7,160.36} & \textbf{0.453} & \textbf{1.280} & \textbf{0.017} & \textbf{1.751} \\
\midrule
Direct Qwen3-VL-8B & 42.11 & 11,975.45 & 0.696 & 3.319 & 0.019 & 4.033 \\
\rowcolor{black!6}\textbf{+FocusEarth} & \textbf{55.91} & \textbf{5,510.47} & \textbf{0.519} & \textbf{1.521} & \textbf{0.018} & \textbf{2.058} \\
\bottomrule
\end{tabularx}
\vspace{-2mm}
\end{table}

We evaluate inference efficiency on all 3,738
questions of MME-RealWorld-RS using an NVIDIA
RTX 5090 GPU (\tabref{tab:cost}).
Answering-stage latency is measured after 60
warm-up calls using precomputed region selections.
With Qwen3-VL-8B, FocusEarth attains an overall
accuracy of 55.91\%, 13.80 percentage points
above the direct-inference baseline, while
reducing visual tokens and answering-stage
latency by 54.0\% and 49.0\%, respectively.
For Qwen2.5-VL-3B, the corresponding reductions
are 54.3\% and 40.1\%, respectively.
Most of the latency savings occur before the
first token is generated, with the TTFT of
Qwen3-VL-8B decreasing from 3.319 to 1.521 s.
Subsequent answer-generation time remains
nearly unchanged for both models.
The separately measured scoring time across
the four image--text encoders averages
0.385 s per question.

The visual-token reduction arises from both
full-scene resizing and region-protected
token compression.
For Qwen3-VL-8B, the resized full-scene image
yields an average of 7,843.88 post-merger
visual tokens, which are further reduced to
5,510.47 through region-protected compression,
a reduction of 29.7\%.
Operating on post-merger features, region-protected
compression reduces the visual sequence length
processed by the language model rather than the
input to the visual encoder.

\begin{figure}[!t]
\centering
\includegraphics[width=\columnwidth]{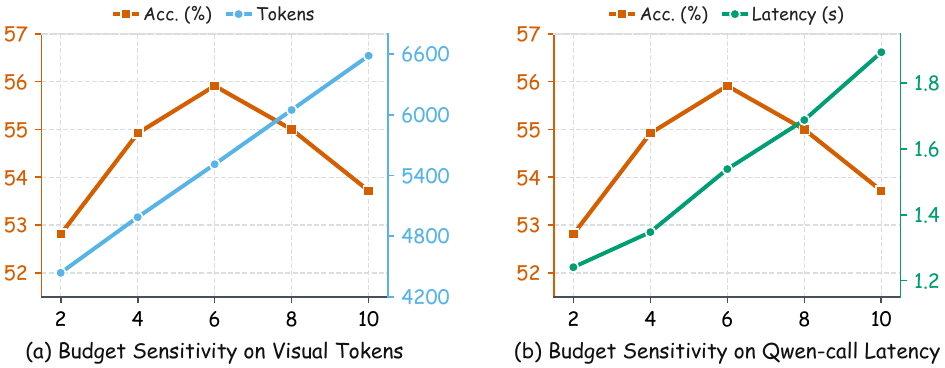}
\caption{Sensitivity of answer accuracy,
visual-token count, and Qwen-call latency to
the region budget $K$ on MME-RealWorld-RS.
Latency excludes input preparation and
offline region scoring and selection.}
\label{fig:budget_sensitivity}
\vspace{-2mm}
\end{figure}

\figref{fig:budget_sensitivity} shows how the
region budget $K$ affects answer accuracy,
visual-token count, and Qwen-call latency
on MME-RealWorld-RS.
Accuracy increases from 52.81\% at $K=2$
to 55.91\% at the default budget of $K=6$,
before declining to 53.72\% at $K=10$.
Larger region budgets are associated with
higher visual-token costs and longer
Qwen-call latency.
Increasing the region budget expands the
allocation for individually retained features,
but does not necessarily improve answer accuracy.

\section{Limitations}
\label{sec:limitations}

\begin{figure}[!t]
\centering
\includegraphics[width=\columnwidth]{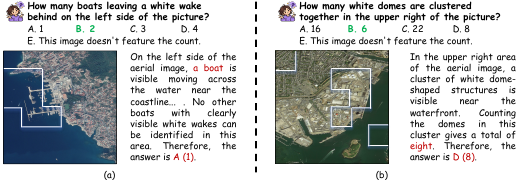}
\caption{Illustrative counting failures on MME-RealWorld-RS: (a) boats leaving white wakes, (b) clustered white domes.}
\label{fig:counting_failures}
\end{figure}

Counting remains a challenging task for FocusEarth.
On MME-RealWorld-RS, its counting accuracy remains
below that of ZoomEarth and WeaveEarth
(\tabref{tab:mme_results}).
Similar difficulties are observed on LRS-VQA
and XLRS-Bench-Lite.
The failure cases in \figref{fig:counting_failures}
illustrate the difficulty of accurately enumerating
small and densely distributed objects.

The fixed candidate grid and region budget may
limit the coverage of spatially dispersed instances
or objects spanning multiple regions.
Moreover, feature protection operates on the
encoded representation of the resized full-scene
image and cannot recover fine-grained details
lost during resizing or visual encoding.
Spatial aggregation of unprotected features may
further reduce the distinguishability of nearby
small objects.
Future work will explore adaptive region selection
and finer-grained feature preservation to improve
instance-level perception under limited visual-token
budgets.

\section{Conclusion}
\label{sec:conclusion}

We presented FocusEarth, a training-free framework
that uses question-guided region selection to direct
feature-level compression while preserving the spatial
layout of the full scene. Across four UHR RS-VQA
benchmarks, FocusEarth achieves accuracies of 55.91\%,
34.32\%, 52.99\%, and 62.85\%, respectively.
With Qwen3-VL-8B, it reduces the visual tokens delivered
to the language model by 54.0\% and answering-stage
latency by 49.0\% relative to direct inference, with
region scoring and selection precomputed offline.
Under matched visual-token budgets, controlled
experiments show that aligning feature protection
with selected source regions improves answer accuracy.

{\renewcommand{\baselinestretch}{0.95}\selectfont
\bibliographystyle{IEEEtran}
\bibliography{refs}}
\end{document}